\documentclass{article}

\usepackage{PRIMEarxiv}
\usepackage{natbib}

\usepackage{amsmath}
\usepackage{amssymb}
\usepackage{mathtools}
\usepackage{amsthm}

\theoremstyle{plain}

\newtheorem{proposhttps://www.overleaf.com/project/61edf01cc636963f7bdd0b9cition}[theorem]{Proposition}

\theoremstyle{definition}

\theoremstyle{remark}

\long\def\cut#1{}

\usepackage[utf8]{inputenc} 
\usepackage[T1]{fontenc}    
\usepackage[hidelinks]{hyperref}       
\usepackage{url}            
\usepackage{booktabs}       
\usepackage{amsfonts}       
\usepackage{nicefrac}       
\usepackage{microtype}      
\usepackage{xcolor}         

\usepackage{graphicx}
\usepackage{soul}

\newcommand{\x}{\mathbf{x}}
\newcommand{\real}{\mathbb{R}}

\newcommand{\ourmodel}{{\it fv-BNN}}
\newcommand{\ourmodels}{{\it fv-BNNs}}
\newcommand{\eg}{\textit{e.\,g., }}

\newcommand{\ie}{\textit{i.\,e., }}

\newcommand{\pbnn}{p_{_{\text{BNN}}}}
\newcommand{\qbnn}{q_{_{\text{BNN}}}}
\newcommand{\pfv}{p_{{\text{fv}}}}
\newcommand{\sfiprior}{SFI prior}
\newcommand{\constprior}{non-functional prior}
\newcommand{\rosettaprior}{stability prior}

\author{%
  Hunter Nisonoff \\
  Center for Computational Biology\\
  University of California, Berkeley\\
  Berkeley, CA 94720\\
  \texttt{hunter\_nisonoff@berkeley.edu} \\
  \And
  Yixin Wang \\
  Department of Statistics\\
  University of Michigan\\
  Ann Arbor, MI 48109\\
  \texttt{yixinw@umich.edu} \\
  \And
  Jennifer Listgarten\\
  Department of Electrical Engineering \& Computer Sciences\\
  Center for Computational Biology\\
  University of California, Berkeley\\
  Berkeley, CA 94720\\
  \texttt{jennl@berkeley.edu}
}
\title{Augmenting Neural Networks with Priors on Function Values}

\begin{document}
\maketitle

\begin{abstract}
  The need for function estimation in label-limited settings is common in the natural sciences. At the same time, prior beliefs of function values are often available in these domains. For example, data-free biophysics-based models can be informative on protein properties, while quantum-based computations can be informative on small molecule properties. How can we coherently leverage such prior beliefs to help improve a neural network model that is quite accurate in some regions of input space---typically near the training data---but wildly wrong in other regions?  Bayesian neural networks (BNN) enable the user to specify prior information only on the neural network weights, not directly on the function values. Moreover, there is in general no clear mapping between these. Herein, we tackle this problem by developing an approach to augment BNNs with prior information on the function values themselves.
  Our probabilistic approach yields predictions that naturally rely more heavily on the prior information when the BNN epistemic uncertainty is large, and more heavily on the neural network when the epistemic uncertainty is small.
\end{abstract}

\section{Incorporating domain knowledge into neural network models}
\label{introduction}
Effective application of supervised machine learning---especially in label-limited settings---relies on the ability to incorporate domain knowledge. Computer vision, for example, has greatly benefited
from incorporating translation equivariance through the convolution operator, in the form of Convolutional Neural Networks. Originally tackled by data augmentation strategies, increasingly, broader and broader classes of equivariances, such as rotational equivariance suitable for molecules, are instead now formally incorporated as constraints into neural networks. Similarly, architectural inductive biases such as attention, residual connections, dropout, and so forth, have greatly benefited a wide range of application areas.  However, relatively little effort has been made to directly incorporate prior knowledge of the function values themselves, such as is often available in many scientific domains, such as biology, chemistry, material science, and beyond.

In these settings, function estimation has traditionally been performed by data-free physics-based modeling, such as biophysics-based models of proteins, quantum-based models of molecules and materials, and so forth. Increasingly, as laboratory techniques for label measurement improve in cost and scale, machine learning-based predictive modeling is starting to supplant these data-free approaches. In particular, on test points similar to the training data, the machine learning models may be more accurate than their data-free counterparts which rely on approximations of the underlying physics for computational tractability. However, the number of labeled data in these domains is often minuscule compared to the size of the input space for which one seeks to make predictions. Thus, traditional data-free models are typically more accurate as test points become less similar to the training data. In particular, we generally expect the data-free models to have roughly comparable accuracy throughout the input space. Consequently, it stands to reason that coherently blending these two modeling approaches in a manner that appropriately navigates their strengths and weaknesses may give us the best of both worlds. 
Herein, we do so by carefully constructing functional prior distributions that depend in part on the data-free models.


\begin{figure*}[t!]
\vskip 0.2in
\begin{center}
\centerline{\includegraphics[width=0.8\linewidth]{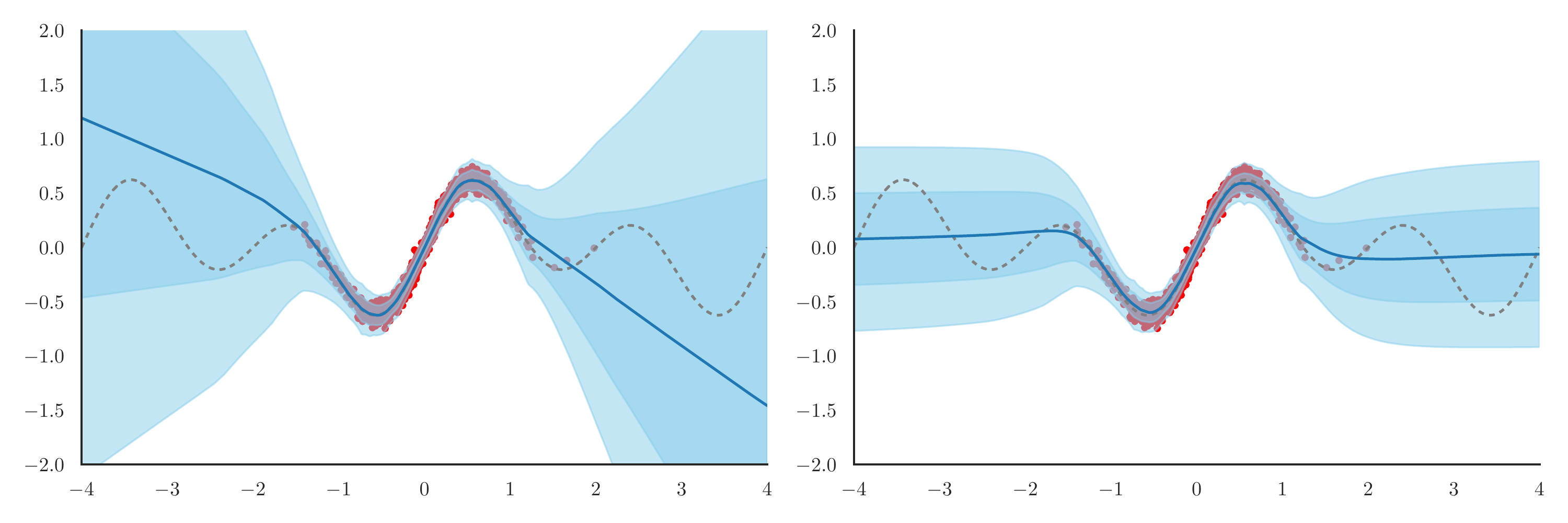}}
\caption{Illustration of \ourmodel~on a one-dimensional, synthetic regression task. The horizontal axis shows the value of the single feature, and the vertical axis shows the regression target values. The left shows a standard Laplace-approximated BNN; the right shows our \ourmodel\ that augments the standard BNN with a zero-mean function-value prior applied the BNN. The dashed grey line corresponds to the true function; red dots show the training data; the dark blue line shows the posterior predictive mean of the BNN, and the dark and light-shaded blue regions show respectively the first and second standard deviations of the posterior predictive distribution.}
\label{fig:toy-example}
\end{center}
\vskip -0.2in
\end{figure*}

Prior functional beliefs can be thought of as having two main components: one that specifies the smoothness properties of the function, and another that specifies beliefs about function values themselves. In Gaussian process (GP) priors~\citep{rasmussen2006GP}, these two types of information are specified, respectively, by the covariance kernel function, and the mean function, although the latter is often set to zero. In contrast,  for Bayesian Neural Networks (BNNs), the prior is on the network weight parameters, indirectly implying a functional prior, and thereby entangling the two components in a way that cannot be readily understood or specified by the practitioner. Now suppose one has access to a prior which specifies only the function values (\eg in the form of a single protein stability belief for each protein), but not anything about the smoothness of the function. What is an effective way to perform regression in such a scenario? On the one hand, one might consider using a GP prior with an informative mean function. However, it can be challenging to engineer a suitable GP kernel and scale these approaches. Moreover, neural networks are often the solution of choice owing to their generally strong performance across many tasks. Herein, we focus on neural network models for these reasons, and because we have found them to work well on our scientific problems of interest. However, in relying on neural networks, a technical challenge of how to coherently incorporate prior beliefs about function values arises. For example, suppose your prior beliefs on function values come from a biophysics-based model that itself makes a prediction at each point of the input space. In general, there is no mapping to convert such beliefs into a prior over the weight space of a neural network.

We tackle this problem of incorporating function value beliefs into BNNs by developing and testing {\it function-value-prior-augmented-Bayesian Neural Networks} (\ourmodels). 
Although recent progress has been made in developing functional priors (\ie~stochastic process priors) for neural networks~\citep{flam2017mapping,flam2018characterizing,matsubara2021ridgelet, sun2018functional}, as we discuss in Section~\ref{sec:related-work}, these approaches are not readily applicable to our problem.
Our modeling approach enables us to modulate the errors that BNNs tend to make far from the training data, as 
illustrated in Figure \ref{fig:toy-example}. Therein, the
 BNN makes egregious mean predictive errors as one moves away from the training data, but incorporating a simple zero-mean prior through our approach modulates the predictive distribution to be more accurate. As we show in our experiments on real data, even such a simple zero-mean prior can help improve protein property prediction because it encodes the knowledge that proteins mutated substantially away from a naturally occurring protein are unlikely to stably fold~\citep{biswas2020low}. We, however, also demonstrate effective use of more informative priors.


Our approach has the desired property of relying more on the prior over function values when the standard (non-augmented) BNN has large epistemic uncertainty (\ie~that arising from lack of knowledge because of lack of data), while relying more on the non-augmented BNN when the epistemic uncertainty is smaller. Although the topic of effectively capturing epistemic uncertainty remains an active area of research~\citep{pmlr-v139-izmailov21a}, in our experiments, we found that existing BNN methods have done so well enough for us to usefully incorporate prior beliefs over function values on real scientific data sets. As the estimation of epistemic uncertainty improves, our approach will correspondingly benefit. As with any use of prior beliefs, our proposed methodology is useful only to the extent that the prior information is useful for the problem at hand. As we show in our experiments, we have found both simple (constant-valued) and rich (biophysics-based) priors that improve prediction on our real data sets.



\section{Related work}
\label{sec:related-work}
Techniques to impose monotonicity constraints for function estimation on neural networks span several decades~\citep{archer1993monotonic,sill1998monotonic,muralidhar2018incorp}. More recently, in Sparse Epistatic Networks~\citep{aghazadeh2021epistatic}, compressed sensing techniques were used to impose sparsity in the Fourier basis of the function expressed.
\citet{wilson2016deep} combine function-value priors with neural networks through Deep Kernel Learning (DKL); however, it has been shown that DKL models can exhibit pathological behavior, resulting in poor uncertainty quantification, and therefore, may not reliably incorporate prior beliefs~\citep{ober2021promises}. As progress is made on DKL, this approach may become more feasible.

\citet{fortuin2022priors} provides a thorough overview of priors used for BNNs, of which we now discuss those most relevant to our work.
\citet{NEURIPS2020_95c7dfc5} introduce a method to incorporate equality, inequality, and logical constraints to BNNs through the introduction of user-specified constraint regions.
{\citet{flam2017mapping,flam2018characterizing}}, and {\citet{matsubara2021ridgelet}} learn approximate mappings from a GP prior to a prior in weight space by way of a variational approximations, or through ridgelet transforms. To the extent that the approximations therein become exact, the resulting methods are implementing standard GP prior regression, albeit with potential computational scaling benefits. However, it's interesting to ponder if the approximation gaps themselves may be conferring some benefits that are not well-understood.
  \citet{sun2018functional} generalize this type of modeling to any stochastic process prior amenable to sampling, not just that of a GP.
  There are two challenges to using such an approach for our intended purposes. The first is the need to specify a useful stochastic process prior, and the second is the requirement to be able to sample from it.
  The first challenge is difficult because the type of function-value prior information we work with does not specify a unique stochastic process prior, leading us to develop a new type of stochastic process prior; moreover, it turns out that this prior cannot be sampled. These ideas will be made evident in Section~\ref{methods}, but briefly, the issue is that we are only provided with prior beliefs about the function values, and not about function smoothness properties, which we, therefore, need to obtain in some general manner. We will obtain the smoothness portion of the stochastic process prior implicitly from the BNN that we are augmenting with our approach. The resulting prior is not amenable to sampling, a key requirement for~\citet{sun2018functional}.


\citet{hafner2020priors} propose a method that augments the original BNN training objective with an additional variational objective that encourages the BNN to output high epistemic uncertainty away from the training data. While there may be ways to adapt this approach to incorporating function value priors, it would require knowing the test set ahead of time, whereas our proposed method does not require knowing the test set ahead of time. 

\citet{osband2018randomized} and \citet{dwaracherla2022ensembles} combine random prior functions with Deep Ensembles in order to incorporate prior beliefs away from the training data. A major drawback of this approach is that it requires the use of Deep Ensembles, whereas our approach can be readily used with any (approximate) Bayesian inference method.

\section{Function-value-prior-augmented-Bayesian Neural Networks}
\label{methods}

We assume that we are given training data, 
\mbox{$D=\{ ( \mathbf{x}_i, y_i )\}_{i=1}^{N}$},
with inputs, $\mathbf{x}_i\in\real^{d}$, and regression labels, $y_i\in\real$, that have been generated according to
\begin{align}
\label{eq:data_generation}
y_i = f({\x}_i) + \epsilon_i, \quad \epsilon_i \sim \mathcal{N}(0, \sigma_0^2),
\end{align}
for some arbitrary and unknown function, $f(\cdot)$, and with $\sigma^2$ unknown. 
We assume that $f(\cdot)$ can be captured by a neural network, $f_{\textit{NN}}(\theta)$ with unknown weight parameters,
$\theta\in\real^{m}$. It is with respect to this neural network that we will be Bayesian, using any available standard BNN approach of our likings, such as a Laplace approximation~\citep{daxberger2021laplace}, or Deep Ensembles~\citep{lakshminarayanan2016simple}; these leverage a prior on the neural network weights, not the function values. As detailed shortly, we augment this BNN with our prior beliefs on function values, yielding our proposed approach, \mbox{\ourmodel}. 
%

\subsection{Neural network weight priors} 
We start our exposition with the standard BNN approach, wherein a prior, $p(\theta)$, is placed over the neural network weights; this gets combined with the likelihood, $p(D \mid \theta)$, to obtain the posterior parameter distribution, $p(\theta \mid D)$. Predictions for a test point, $\x\in\real^{d}$, are given by the posterior predictive distribution,
\begin{align*}
  p(y \mid \x, D) = \int_\Theta p(y \mid \theta, \x) p(\theta \mid D) d\theta.
\end{align*}

\subsection{Function-value priors} 

Our function-value prior beliefs are assumed to be given, for each 
point\footnote{This setting can readily be adapted to not require domain knowledge over the entire domain, by effectively setting the variance for parts of the domain to infinity.}
in the function domain,
$\x$, by a mean and variance in a Gaussian distribution,  ${\pfv(f(\x)) = \mathcal{N}(\mu_{\text{fv}}(\x), \sigma_{\text{fv}}^2(\x))}$.
The variance controls the strength of our prior beliefs, 
which may be known ahead of time, but that we herein treat as a single, scalar hyperparameter, yielding an empirical Bayes approach. To leverage this prior information with our BNN, we will want to lift the pointwise information into a prior over functions, which can be accomplished by writing it in the form of a diagonal GP prior on the function space,
\begin{align*}
  \pfv(f) = \mathcal{GP}\Big(\mu_{\text{fv}}(\x), \text{diag}\big(\sigma_{\text{fv}}^2(\x)\big)\Big).
\end{align*}
It's important to emphasize that unlike in standard GP regression, our GP prior is not intended to encode any smoothness properties of the function---hence its diagonal kernel---only to capture the entirety of our function-value prior knowledge, and pointwise uncertainty we may have about these. We will rely on the standard BNN prior to provide smoothness information.

\subsection{Combining priors over weights and function values} 

We now have two priors, one from the standard BNN approach over the weights of the neural network, $p(\theta)$, and our newly introduced prior over function values, $\pfv(f)$, each over different random variables that define the function, thereby making it challenging to coherently combine them. However, note that the prior on the neural network weights, $p(\theta)$, implies a prior on the neural network function, $\pbnn (f)$. Although we do not know the expression for this implicit functional prior, it in turn implies a posterior over functions, $\pbnn (f \mid D)$, arising from just the standard (non-augmented) BNN. This implicit posterior can be written according to Bayes rule as follows,
\begin{align*}
\pbnn (f \mid D) \propto \pbnn (f) p(D \mid f),
\end{align*}
where $p(D \mid f)$ is the corresponding implicit likelihood. Now that we have two priors, $\pfv(f)$ and $\pbnn(f)$, in terms of the same random variable, $f$, we can ask: how should they be meaningfully combined to tackle our problem of interest? 
%

%

First, it will be convenient to approximate the BNN posterior distribution, $\pbnn(f \mid D)$, with a variational approximation in the form of a Gaussian process with a diagonal covariance matrix, $\qbnn(f \mid D)$. Although this family of approximations could be overly restrictive in some problem settings, in ours, the limitations are minor compared to its enabling of closed-form computation that is cheap to compute. Morever, it works well empirically. We discuss this issue further at the end of the section.

Our variational approximation is derived from minimizing the ``inclusive" KL divergence,
$KL(p \, || \, q)$, that is, in the opposite direction from what is typically used
in variational inference; this divergence is sometimes used because it more accurately captures posterior uncertainty, even though it is typically more difficult to optimize for~\citep{naesseth2020inclusive}. However, in our setting,
we use it because it is easier to work with. Specifically, it enables us to fit the variational posterior by moment matching at each point, $\x$, to get the pointwise posterior mean and diagonal covariance,
%
 \mbox{$q_{_{\text{BNN}}}(f(\x) \mid D) = \mathcal{N}(\mu_{_{\text{BNN}}}(\x), \sigma_{_{\text{BNN}}}^2(\x))$}, where,
  \mbox{$\mu_{_{\text{BNN}}}(\x) = \mathbb{E}_{\pbnn(f \mid D)}\big[f(\x) \big]$} and 
  $ {\text{diag}\big(\sigma_{_{\text{BNN}}}(\x) \big) = \text{diag}\big(\text{Var}_{\pbnn(f \mid D)}\big[f(\x) \big]\big)}$
%
are straightforward to estimate for many approximate Bayesian inference procedures, such as  Laplace and Deep Ensembles. ``Lifting'' these pointwise posteriors into one posterior over functions, we obtain the following GP posterior,
\begin{align}
  q_{_{\text{BNN}}}(f \mid D) 
               &=\mathcal{GP}\Big(\mu_{_{\text{BNN}}}(\x), \text{diag}\big(\sigma_{_{\text{BNN}}}^2(\x)\big)\Big),
               \label{eq:moment-matching-posterior}
\end{align}
which conveniently produces the same pointwise posterior predictive mean and variance predictions as the BNN posterior it approximates, $\pbnn(f \mid D)$. Moreover, these are commonly-used quantities for downstream use from BNNs~\citep{amini2020deep,lakshminarayanan2016simple}. 

Because our approximate posterior, ${\qbnn(f \mid D)}$, is a GP with a diagonal covariance matrix and our likelihood is Normal, it follows from Bayes rule, that the corresponding prior distribution, $\qbnn(f)$, is also a GP with a diagonal covariance matrix. 

We are now ready to turn to the question of how to construct a new prior, $p(f)$, that better reflects our function-value beliefs. Our motivating intuition is that we would like to modulate the BNN prior according to how well these functions satisfy our function-value beliefs. We accomplish this by defining a new prior, our combined {\it function-value modulating product prior},
\begin{align}\label{eq:productprior}
  p(f) \propto \pfv(f) \, \qbnn(f),
\end{align}
which requires some explanation as we have abused the notation of probability densities to describe a "product" of stochastic processes. In general, it is not clear how to define a product of stochastic processes that yields another stochastic process (\ie staisfying the consistency constraints required by the Kolmogorov extension theorem). However, for two GPs with diagonal covariances, a natural such definition exists. Given two GPs with diagonal covariances, $p_1(f)$ and $p_2(f)$, we can define their product, $p_{12}(f)$ as the following stochastic process:
 \begin{align*}
  p_{12}(f) &\propto p_1(f) \, p_2(f)\\
       &\propto  \mathcal{GP}\Big(\mu_{1}(\x), \text{diag}\big(\sigma_{1}^2(\x)\big)\Big)
  \\
       &\qquad\times \mathcal{GP}\Big(\mu_{2}(\x), \text{diag}\big(\sigma_{2}^2(\x)\big)\Big)\\
       &:= \mathcal{GP}\Big(\mu_{12}(\x), \text{diag}\big(\sigma_{12}^2(\x)\big)\Big),
\end{align*}
where,
\begin{align*}
  \mu_{12}(\x) &= \frac{\sigma_{1}^2(\x)^{-1}\, \mu_{1}(\x) + \sigma_{2}^2(\x)^{-1}\mu_{2}(\x)}
  {{\sigma_{1}^2(\x)^{-1}} + {\sigma_{2}^2(\x)}^{-1}},
  \\
  \sigma_{12}^2(\x) &= \Big({\sigma_{1}^2(\x)^{-1}} + {\sigma_{2}^2(\x)}^{-1}\Big)^{-1}.
\end{align*}

We will now show how this {\it function-value modulating product prior} yields a posterior that encodes our original modeling goals. In particular, the posterior can be written as,
\begin{align}   
  p(f \mid D) &\propto p(f) \, p(D \mid f)\\
          &\propto \pfv(f) \, \qbnn(f) \, p(D \mid f)\\
              &\propto \pfv(f) \, \qbnn(f \mid  D).\label{eqn:1}\\
                           &= 
           \mathcal{GP}\Big(\mu_{\text{fv}}(\x), \text{diag}\big(\sigma_{\text{fv}}^2(\x)\big)\Big)
           \\
           &\qquad\times \mathcal{GP}\Big(\mu_{_{\text{BNN}}}(\x), \text{diag}\big(\sigma_{_{\text{BNN}}}^2(\x)\big)\Big)\\
           &= \mathcal{GP}\Big(\mu(\x), \text{diag}\big(\sigma^2(\x)\big)\Big),
\end{align}
where,
\begin{align}
  \mu(\x) &= \frac{\sigma_{_{\text{BNN}}}^2(\x)^{-1}\, \mu_{_{\text{BNN}}}(\x) + \sigma_{{\text{fv}}}^2(\x)^{-1}\mu_{_{\text{fv}}}(\x)}
  {{\sigma_{_{\text{BNN}}}^2(\x)^{-1}} + {\sigma_{_{\text{fv}}}^2(\x)}^{-1}},\label{eqn:fvbnn-mean}
  \\
  \sigma^2(\x) &= \Big({\sigma_{_{\text{BNN}}}^2(\x)^{-1}} + {\sigma_{_{\text{fv}}}^2(\x)}^{-1}\Big)^{-1},
\label{eqn:fvbnn-var}
\end{align}
for which all the terms are now readily computable.
Equations \ref{eqn:fvbnn-mean} and~\ref{eqn:fvbnn-var} enjoy an intuitive interpretation: the posterior mean is a convex combination of the original BNN posterior mean and the function-value prior mean, where the weights are determined by the epistemic uncertainty of the original BNN and the strength of the prior. Thus, we can see that the {\it function-value modulating product prior} successfully achieves our goal of incorporating our function-value prior beliefs.

The final posterior predictive distribution for the \ourmodel\ is 
\begin{align*}
    p(y \mid \x , D) = \mathcal{N}(\mu(\x),\ \sigma^2(\x) + \sigma_0^2),
\end{align*}
where $\mu(\x)$ and $\sigma^2(\x)$ are given by Equations
\ref{eqn:fvbnn-mean} and~\ref{eqn:fvbnn-var}, and $\sigma_0^2$, the  noise in the data from Equation \ref{eq:data_generation}, can be estimated from the data, using for example, the mean squared error of validation set predictions. 

One potential drawback of the diagonal covariance structure of the BNN variational posterior is that at test time, correlations between test points are ignored. These correlations are particularly important for batch active learning, whereas our interest is in improving prediction. Although incorporating a full covariance posterior might be expected to improve results beyond those demonstrated herein, this would come at the cost of substantial complexity. Assessing this direction is beyond the scope of  the present work, however, we detail how it could be done in the Appendix.
 We emphasize that our variational approximation requires no hyperparameter tuning, and, as shown in experiments, works well.

\section{Experiments}
\label{experiments}

To understand and characterize the value of our proposed approach,~\ourmodel, we first constructed a synthetic one-dimensional example that could readily be understood at an intuitive level. Then we applied our method to real prediction problems in the natural sciences to empirically evaluate how much it could help improve prediction compared to alternative methods. Next, we go through these experiments in detail. When published, the accompanying code will be made public.

\subsection{Illustrative 1D example}

Our illustrative example (Figure \ref{fig:toy-example}) was constructed by sampling 1000 points from {$y = 0.3  \sin(\frac{\pi}{2}  x) + 0.4 \sin(\pi x)$}, keeping 800 for training and using the remaining 200 for validation. We construct an approximate BNN posterior using the last-layer Laplace approximation suggested by~\citet{kristiadi2020being} using the Laplace PyTorch library~\citep{daxberger2021laplace}. While Deep Ensembles and the Laplace approximations gave similar posterior mean predictions, we found that the Laplace approximation gave much better epistemic uncertainty.

Figure~\ref{fig:toy-example} compares the BNN fit to this data with our \ourmodel\ using a zero-mean prior with a variance equal to the empirical variance of the training and validation data  ${(\sigma=0.43)}$. This function-value prior encodes our prior belief that the function is centered around zero. Whereas the BNN extrapolates erroneously as we move away from the training data, \ourmodel\ regresses back toward the prior mean in regions of high epistemic uncertainty. On the basis of this sanity check, we next moved on to experiments on real data.

\subsection{Experiments on real scientific data}

We chose two protein datasets commonly used for benchmarking protein fitness prediction, wherein the goal is to predict the ``fitness" (\ie scalar property of interest) from the protein sequence of amino acids. This problem is of great interest for a number of reasons including for protein engineering and mutation effect prediction~\citep{wittmann2021informed,aghazadeh2021epistatic,biswas2020low,hsu2022nbt,riesselman2018deep,hopf2017mutation, russ2020evolution}. We also used one small-molecule data set wherein the goal is to predict solubility from some representation of the small molecule, a property frequently required for drug development~\citep{di2012bridging}. We describe the details of our chosen data sets shortly. 

\subsubsection{Approaches compared against}

The basis of all of our experiments on each data set is a single neural network architecture. For the two protein data sets, model architectures and optimization parameters were chosen by cross-validation (see Appendix). The avGFP data set used a fully-connected neural network with two hidden layers, each with 100 dimensions, ReLU non-linearities, and was optimized using Adam~\citep{DBLP:journals/corr/KingmaB14} with a weight-decay of $0.0001$.  The GB1 data set used a  fully-connected neural network with 1 hidden layer containing 300 dimensions, ReLU non-linearities, and was optimized using Adam with a weight-decay of $0.0001$. The small molecule data set used the Graph Convolution Neural Network
(GCNN)~\citep{duvenaud2015convolutional} with the default molecular
featurization and neural network hyperparameters as provided in the
DeepChem library~\citep{Ramsundar-et-al-2019}.

Each data set had its own neural network architecture. From this architecture, we compared the following approaches on all three of our data sets:
\begin{enumerate}
    \item {\it NN}: The neural network with a point estimate of the weights when fit using a cross-entropy loss on the training data using the validation data for early stopping.
    \item {\it BNN}: The neural network is used with Deep Ensembles, using five networks in the ensemble. Each network was trained in the same manner as the NN above, but with different weight initialization as done by~\citet{lakshminarayanan2016simple}. This approach  has been demonstrated to provide good estimates of the predictive posterior
distribution in a range of scenarios~\citep{wilson2020bayesian,pearce2020uncertainty,gustafsson2020evaluating}.

    \item \ourmodel: One or two different priors are used to augment the BNN, as described in each experimental section.
    \item {\it STACKING}: A linear regression model is fit using two features which are predictions made from (i) the BNN and (ii) the prior that was used in~\ourmodel, as described below. The stacker thus had three free parameters which were fit using the validation data with a log-likelihood loss. 
\end{enumerate}
 An important distinction between our approach and the stacking regressor is that
our method is able to use the epistemic uncertainty of the original
BNN to modulate the influence of the prior. Thus, stacking provides a
strong baseline to test whether our approach improves over simply
ensembling the two models: the neural network and the prior.

We do not compare to any of the approaches in the literature whose goal is to approximate a GP process regression with a neural network, as their goals are not to augment a BNN with a prior over function values.

We evaluated the performance of each modeling approach using each of the log-likelihood (LL), the root mean squared error (RMSE), and the mean absolute error (MAE). The statistical significance of the comparison of pairs of methods is provided in the Appendix.

\begin{table*}[t!]
\caption{Results on avGFP fluorescence prediction. Plus-minus indicates the standard error computed from 10 train-test splits of the data.}
\label{table:GFP}
\begin{center}
\begin{small}
\begin{sc}

\begin{tabular}{llll}
\toprule
           Method &   Log Likelihood &            RMSE &             MAE \\
\midrule
               NN & $-3.60 \pm 0.13$ & $1.03 \pm 0.01$ & $0.70 \pm 0.01$ \\
              BNN & $-3.15 \pm 0.12$ & $1.03 \pm 0.01$ & $0.71 \pm 0.01$ \\
 Stacking BNN+\constprior & $-3.93 \pm 0.16$ & $1.04 \pm 0.01$ & $0.72 \pm 0.01$ \\
 Stacking BNN+\rosettaprior & $-3.56 \pm 0.17$ & $0.97 \pm 0.01$ & $0.67 \pm 0.01$ \\
\ourmodel\ (\constprior) & $-1.82 \pm 0.00$ & $0.97 \pm 0.01$ & $0.67 \pm 0.01$ \\
\ourmodel\ (\rosettaprior) & $-1.53 \pm 0.00$ & $0.85 \pm 0.01$ & $0.57 \pm 0.01$ \\
\bottomrule

\end{tabular}


\end{sc}
\end{small}
\end{center}
\vskip -0.1in
\end{table*}

\begin{table*}[t!]
\caption{Results on GB1 binding affinity prediction. Plus-minus indicates the standard error computed from 10 train-test splits of the data.}
\label{table:gb1}
\begin{center}
\begin{small}
\begin{sc}
\begin{tabular}{llll}
\toprule
                       Method &   Log-Likelihood &            RMSE &             MAE \\
\midrule
                           NN & $-1.00 \pm 0.03$ & $0.65 \pm 0.02$ & $0.50 \pm 0.02$ \\
                          BNN & $-0.91 \pm 0.02$ & $0.59 \pm 0.02$ & $0.45 \pm 0.01$ \\
Stacking: BNN+\constprior & $-1.10 \pm 0.05$ & $0.69 \pm 0.02$ & $0.54 \pm 0.02$ \\
     Stacking: BNN+\rosettaprior & $-1.09 \pm 0.06$ & $0.68 \pm 0.03$ & $0.54 \pm 0.02$ \\
           \ourmodel\ (\constprior) & $-0.73 \pm 0.05$ & $0.48 \pm 0.03$ & $0.28 \pm 0.05$ \\
          \ourmodel\ (\rosettaprior) & $-0.57 \pm 0.00$ & $0.38 \pm 0.00$ & $0.13 \pm 0.00$ \\
\bottomrule
\end{tabular}

\end{sc}
\end{small}
\end{center}
\vskip -0.1in
\end{table*}

\begin{table*}[t!]
\caption{Results on aqueous solubility prediction. Plus-minus indicates the standard error computed from 10 train-test splits of the data.}
\label{table:solubility}
\begin{center}
\begin{small}
\begin{sc}

\begin{tabular}{llll}
\toprule
       Method &   Log-Likelihood &            RMSE &             MAE        \\
\midrule
                          NN & $-2.39 \pm 0.04$ & $1.89 \pm 0.03$ & $1.31 \pm 0.02$ \\
                         BNN & $-2.06 \pm 0.03$ & $1.71 \pm 0.02$ & $1.15 \pm 0.01$ \\
     Stacking: BNN+\sfiprior & $-2.35 \pm 0.04$ & $1.01 \pm 0.01$ & $1.13 \pm 0.01$ \\
\ourmodel\ (\sfiprior) & $-2.04 \pm 0.03$ & $1.67 \pm 0.02$ & $1.12 \pm 0.01$  \\

\bottomrule
\end{tabular}

\end{sc}
\end{small}
\end{center}
\vskip -0.1in
\end{table*}

\subsubsection{Priors over function values}

We use three types of priors over function values in our experiments. Two are used for protein fitness prediction, and one is used for solubility prediction. Our goal is not to find or develop the best priors possible for these problems, only to demonstrate that reasonable priors exist, both simple, and rich, and that our proposed method can make use of these in an effective manner, as demonstrated by improved prediction accuracy.

\paragraph{Constant zero-fitness prior.} 
The first prior for protein fitness is a simple prior that takes on the value zero throughout all of protein space, which we denote as our {\it \constprior}. Although this may seem such a simple prior as to be useless, there is substantial evidence that protein fitness landscapes are typically dominated by
non-functional sequences
\citep{kauffman1989nk,arnold2012library,bloom2005thermodynamic,romero2013navigating,wittmann2021informed}. Moreover, there has been substantial work making use of large-scale unsupervised protein data with deep learning to try to learn the generic, non-functional ``holes" in protein space~\citep{biswas2020low}. Indeed, we will find that this~\constprior~is useful to incorporate in our experiments, although not as useful as a much richer source of information based on biophysics, described next.

\paragraph{Stability prior.} Our second prior uses the intuition just described but in a more nuanced manner. In particular, rather than assume that all of protein space is dominated by non-fit proteins, we make the further argument that a major determinant of which proteins are fit, irrespective of the property in question, is their inherent stability.
Consequently, our second protein prior, \rosettaprior, is defined by stability predictions provided by the biophysics-based 
Rosetta~\citep{alford2017rosetta} model. By doing so, we encode our prior beliefs that protein stability is typically a necessary but
not sufficient condition for protein fitness, as well as the beliefs imparted by Rosetta stability estimates. 
It's worth noting that it has previously been
demonstrated that little correlation exists between Rosetta stability predictions and fitness when restricted to proteins predicted to be stable. However, it has also been shown that those proteins predicted to be unstable are more likely to have poor fitness~\citep{gelman2021neural,wittmann2021informed}. 
We converted the real-valued Rosetta scores to a binary ``stable" versus ``not stable", by running a grid search on the predicted stability scores of the combined
training and validation sets to find the Rosetta score that best
separated fit versus not fit sequences according to the
ROC-AUC metric on the relevant prediction property of interest (fluorescence for avGFP or binding activity for GB1). This prior still uses a mean of zero but has a separate variance parameter for the two sets of binary encoded sequences. This allows the model to place a stronger belief that a sequence is not functional if it is determined to be  unstable by the Rosetta score.

\paragraph{Solubility prior.} For the small molecule data set, for which our task was to predict solubility, we used just one prior, our~\sfiprior, which had a mean given by the Solubility Forecast Index (SFI), after scaling and shifting the value to match the units of solubility. The SFI was developed by medicinal chemists to
predict a score correlated to aqueous solubility from physio-chemical
properties of the molecule; it is considered among the most reliable
predictor of aqueous solubility~\citep{hill2010getting}. Our SFI predictions were
computed using~\citep{Walters}.

The~\constprior~and the~\sfiprior~each had a single scalar parameter that needed fitting, reflecting the strength of the prior as a variance as described in the methods. The~\rosettaprior~had two scalar parameters that needed fitting (as in empirical Bayes), corresponding to the strength for stable and unstable proteins as predicted by Rosetta. These parameters were fit with a grid search to
maximize the marginal likelihood on the validation set. 

\subsubsection{Data sets and corresponding experimental set-up}

Next, we provide descriptions of each of our three data sets, and how they were partitioned into train, validation, and test sets. Following these, we discuss the experimental results across all three of these in Section \ref{sec:results}.

\paragraph{Predicting protein fluorescence (avGFP).}
Our first experiment centered on making predictions of green fluorescence on proteins, using a data set comprised of measurements of the naturally occurring
\textit{Aequorea victoria} green fluorescent protein (avGFP), and variants of it up to 15 mutations away, determined by a physical random mutagenesis procedure in the laboratory~\citep{sarkisyan2016local}. There were 51,714 proteins with measured fluorescence in total.
In many protein engineering applications, we are trying to engineer an existing protein to have more of a property (\eg brighter), or to alter its property (\eg change fluorescence wavelength). Often we start with protein sequence
variants close to a naturally existing one---a so-called wild-type (WT) sequence---and predict
fitness values of proteins increasingly further away as we explore the space. To represent this extrapolative scenario of interest, we sampled $3,000$ sequences within two mutations (out of a total of $13,861$) of the avGFP WT sequence, keeping a random $80\%$ of these for training data, and the remaining $20\%$ as our validation set. The test set was composed of all $48,714$ sequences not in the training or validation sets. 
 Rosetta scores for these data
were previously computed by~\citet{gelman2021neural}.

\paragraph{Predicting protein binding affinity (GB1).}
The GB1 landscape~\citep{wu2016adaptation} measures the binding
affinity of all GB1 variants ($4^{20}=160,000)$ at four amino acid
sites to IgG-Fc. Similarly to avGFP, we randomly sampled $500$ sequences within two mutations from the WT sequence, using $80\%$ of these for training, and $20\%$ for validation. The test set comprised all other proteins. We chose $500$ so as to provide a similar fraction of
training sequences from within the pool of variants with two mutations
as was used in the avGFP experiment. However, we found our qualitative conclusions  to be robust to the number of training sequences.
We ensured that half of the sequences used for training and validation were deemed fit,\footnote{Fit was determined by a threshold of 0.5 as suggested by ~\citet{dallago2021flip}.} by sampling appropriately, so as to balance the data set. Rosetta scores were previously computed
by~\citet{wittmann2021informed} using the Triad protein design
software suite (Protabit, Pasadena, CA, USA:
https://triad.protabit.com/) with a Rosetta energy function under the
fixed-backbone protocol, which was found to offer better stability
predictions than the flexible backbone alternative. 

\paragraph{Predicting solubility of drug-like molecules.}
Aqueous solubility is among the most important physical properties
required for a drug molecule, as it facilitates the delivery of the
drug to its target. Accurately predicting aqueous solubility from the
molecular graph has therefore been a key tool in drug discovery. 
SFI does not directly report aqueous solubility, but rather provides a solubility score, which has a very
strong linear relationship to aqueous solubility. As such, we construct
an SFI prior by fitting the SFI scores to solubility measurements on
the training and validation data and using the linearly scaled SFI
predictions as the prior mean.
We used the Therapeutic
Data Commons (TDC) to construct $0.6$/$0.4$ train/test splits using
their default scaffold-split and using $20\%$ of the training points
for validation~\citep{huang2021therapeutics}. 

\subsubsection{Experimental results across all data sets}
\label{sec:results}

There are a number of interesting observations to make from these comparisons between models, across the avGFP fluorescence predictions, the GB1 binding activity predictions, and the solubility prediction (respectively Tables   \ref{table:GFP}, \ref{table:gb1} \ref{table:solubility}). First, the BNN always outperforms the NN by held out log-likelihood, and is either better, or comparable by RMSE and MAE. Second, and of primary interest for our work, the~\ourmodel~with a rich prior, always outperforms the non-augmented BNN, across all three metrics and all three data sets, each with a corresponding p-value no larger than $p=0.002$ (see Appendix). 

\ourmodel~with the simple, constant, zero-fitness prior also  always outperforms the non-augmented BNN, across all three metrics and both protein data sets for which the comparison can be made, each with a corresponding p-value no larger than $p=0.001$ (see Appendix). This underscores how even a simple zero-mean prior on the function values can improve property prediction by encoding the knowledge that proteins mutated away from a naturally occurring sequence are unlikely to be functional. Additionally, with~\ourmodel, the rich prior always outperforms the simple prior; moreover, it provides the best overall performance compared to all other approaches.

\section{Discussion and Conclusions}

Motivated by regression problems in the natural sciences relevant to protein engineering, small molecule, and materials design, and beyond, we have proposed a method, {\it function-value-prior-augmented-Bayesian Neural Networks} (\ourmodels), which enables the user to take prior information on the specific values of the function---such as might be obtained from a biophysical model, or more coarse-grained information---and coherently integrate it into a Bayesian Neural Network setting, for any BNN inference techniques from which the pointwise posterior predictive mean and variance can be extracted or approximated. 

The degree by which our method may outperform a non-augmented BNN depends both on the quality of the prior information itself and its relevance to the task, but also on the extent to which the epistemic uncertainty of the non-augmented BNN is calibrated. From our results on real data, it is clear that each of these criteria can be satisfied sufficiently to observe an improvement over non-augmented BNNs. It might be worth exploring further developments to our strategy that enable specific function smoothness prior information to also be accounted for, when available. 

It's worth noting that the empirical Bayes method we used leveraged a validation set that was sampled in the same manner as the training data set, both of which were distinct distributions from the test set, so as to make for a difficult, extrapolative setting. An interesting area for further investigation could be to explore different strategies for making the validation set different from the training data set in a way that may more closely mimic the test use cases. For example, one could re-partition the training-plus-validation used to more closely mimic an extrapolation to higher-order protein mutations. Indeed, a promising future application of this method will be to leverage it within an {\it in silico} design cycle, such as presented in~\citet{biswas2020low, brookes2019conditioning,fannjiang2020autofocused, madani2021deep, wittmann2021informed}. 

Finally, it would be interesting to apply~\ourmodel~as a surrogate in Bayesian optimization~\citep{snoek2015scalable},  where the ability to place informative function-value priors on the surrogate may allow optimization to be performed with fewer ground-truth function evaluations.


\newpage
\bibliography{references}

\begin{thebibliography}{51}
\providecommand{\natexlab}[1]{#1}
\providecommand{\url}[1]{\texttt{#1}}
\expandafter\ifx\csname urlstyle\endcsname\relax
  \providecommand{\doi}[1]{doi: #1}\else
  \providecommand{\doi}{doi: \begingroup \urlstyle{rm}\Url}\fi

\bibitem[Aghazadeh et~al.(2021)Aghazadeh, Nisonoff, Ocal, Brookes, Huang,
  Koyluoglu, Listgarten, and Ramchandran]{aghazadeh2021epistatic}
Amirali Aghazadeh, Hunter Nisonoff, Orhan Ocal, David~H. Brookes, Yijie Huang,
  O.~Ozan Koyluoglu, Jennifer Listgarten, and Kannan Ramchandran.
\newblock Epistatic {N}et allows the sparse spectral regularization of deep
  neural networks for inferring fitness functions.
\newblock \emph{Nature Communications}, 12\penalty0 (1), 2021.

\bibitem[Alford et~al.(2017)Alford, Leaver-Fay, Jeliazkov, O’Meara, DiMaio,
  Park, Shapovalov, Renfrew, Mulligan, Kappel, et~al.]{alford2017rosetta}
Rebecca~F Alford, Andrew Leaver-Fay, Jeliazko~R Jeliazkov, Matthew~J O’Meara,
  Frank~P DiMaio, Hahnbeom Park, Maxim~V Shapovalov, P~Douglas Renfrew,
  Vikram~K Mulligan, Kalli Kappel, et~al.
\newblock The rosetta all-atom energy function for macromolecular modeling and
  design.
\newblock \emph{Journal of chemical theory and computation}, 13\penalty0
  (6):\penalty0 3031--3048, 2017.

\bibitem[Amini et~al.(2020)Amini, Schwarting, Soleimany, and
  Rus]{amini2020deep}
Alexander Amini, Wilko Schwarting, Ava Soleimany, and Daniela Rus.
\newblock Deep evidential regression.
\newblock \emph{Advances in Neural Information Processing Systems},
  33:\penalty0 14927--14937, 2020.

\bibitem[Archer and Wang(1993)]{archer1993monotonic}
Norman~P. Archer and Shouhong Wang.
\newblock Application of the back propagation neural network algorithm with
  monotonicity constraints for two-group classification problems*.
\newblock \emph{Decision Sciences}, 24\penalty0 (1):\penalty0 60--75, 1993.
\newblock \doi{https://doi.org/10.1111/j.1540-5915.1993.tb00462.x}.

\bibitem[Arnold(2012)]{arnold2012library}
Frances~H Arnold.
\newblock The library of maynard-smith: My search for meaning in the protein
  universe.
\newblock \emph{Microbes and Evolution: The World That Darwin Never Saw}, pages
  203--208, 2012.

\bibitem[Biswas et~al.(2021)Biswas, Khimulya, Alley, Esvelt, and
  Church]{biswas2020low}
Surojit Biswas, Grigory Khimulya, Ethan~C Alley, Kevin~M Esvelt, and George~M
  Church.
\newblock {Low-N} protein engineering with data-efficient deep learning.
\newblock \emph{Nature Methods}, 18\penalty0 (4):\penalty0 389--–396, 2021.

\bibitem[Bloom et~al.(2005)Bloom, Silberg, Wilke, Drummond, Adami, and
  Arnold]{bloom2005thermodynamic}
Jesse~D Bloom, Jonathan~J Silberg, Claus~O Wilke, D~Allan Drummond, Christoph
  Adami, and Frances~H Arnold.
\newblock Thermodynamic prediction of protein neutrality.
\newblock \emph{Proceedings of the National Academy of Sciences}, 102\penalty0
  (3):\penalty0 606--611, 2005.

\bibitem[Brookes et~al.(2019)Brookes, Park, and
  Listgarten]{brookes2019conditioning}
David Brookes, Hahnbeom Park, and Jennifer Listgarten.
\newblock Conditioning by adaptive sampling for robust design.
\newblock In \emph{International Conference on Machine Learning}, pages
  773--782. PMLR, 2019.

\bibitem[Dallago et~al.(2021)Dallago, Mou, Johnston, Wittmann, Bhattacharya,
  Goldman, Madani, and Yang]{dallago2021flip}
Christian Dallago, Jody Mou, Kadina~E Johnston, Bruce~J Wittmann, Nicholas
  Bhattacharya, Samuel Goldman, Ali Madani, and Kevin~K Yang.
\newblock Flip: Benchmark tasks in fitness landscape inference for proteins.
\newblock 2021.

\bibitem[Daxberger et~al.(2021)Daxberger, Kristiadi, Immer, Eschenhagen, Bauer,
  and Hennig]{daxberger2021laplace}
Erik Daxberger, Agustinus Kristiadi, Alexander Immer, Runa Eschenhagen,
  Matthias Bauer, and Philipp Hennig.
\newblock Laplace redux - effortless bayesian deep learning.
\newblock In A.~Beygelzimer, Y.~Dauphin, P.~Liang, and J.~Wortman Vaughan,
  editors, \emph{Advances in Neural Information Processing Systems}, 2021.

\bibitem[Di et~al.(2012)Di, Fish, and Mano]{di2012bridging}
Li~Di, Paul~V Fish, and Takashi Mano.
\newblock Bridging solubility between drug discovery and development.
\newblock \emph{Drug discovery today}, 17\penalty0 (9-10):\penalty0 486--495,
  2012.

\bibitem[Duvenaud et~al.(2015)Duvenaud, Maclaurin, Aguilera-Iparraguirre,
  G{\'o}mez-Bombarelli, Hirzel, Aspuru-Guzik, and
  Adams]{duvenaud2015convolutional}
David Duvenaud, Dougal Maclaurin, Jorge Aguilera-Iparraguirre, Rafael
  G{\'o}mez-Bombarelli, Timothy Hirzel, Al{\'a}n Aspuru-Guzik, and Ryan~P
  Adams.
\newblock Convolutional networks on graphs for learning molecular fingerprints.
\newblock \emph{arXiv preprint arXiv:1509.09292}, 2015.

\bibitem[Dwaracherla et~al.(2022)Dwaracherla, Wen, Osband, Lu, Asghari, and
  Van~Roy]{dwaracherla2022ensembles}
Vikranth Dwaracherla, Zheng Wen, Ian Osband, Xiuyuan Lu, Seyed~Mohammad
  Asghari, and Benjamin Van~Roy.
\newblock Ensembles for uncertainty estimation: Benefits of prior functions and
  bootstrapping.
\newblock \emph{arXiv preprint arXiv:2206.03633}, 2022.

\bibitem[Fannjiang and Listgarten(2020)]{fannjiang2020autofocused}
Clara Fannjiang and Jennifer Listgarten.
\newblock Autofocused oracles for model-based design.
\newblock \emph{Advances in Neural Information Processing Systems}, 33, 2020.

\bibitem[Flam-Shepherd et~al.(2017)Flam-Shepherd, Requeima, and
  Duvenaud]{flam2017mapping}
Daniel Flam-Shepherd, James Requeima, and David Duvenaud.
\newblock Mapping {G}aussian {P}rocess priors to {B}ayesian {N}eural
  {N}etworks.
\newblock In \emph{NIPS Bayesian Deep Learning Workshop}, 2017.

\bibitem[Flam-Shepherd et~al.(2018)Flam-Shepherd, Requeima, and
  Duvenaud]{flam2018characterizing}
Daniel Flam-Shepherd, James Requeima, and David Duvenaud.
\newblock Characterizing and warping the function space of bayesian neural
  networks.
\newblock In \emph{NeurIPS Bayesian deep learning workshop}, 2018.

\bibitem[Fortuin(2022)]{fortuin2022priors}
Vincent Fortuin.
\newblock Priors in bayesian deep learning: A review.
\newblock \emph{International Statistical Review}, 2022.

\bibitem[Gelman et~al.(2021)Gelman, Fahlberg, Heinzelman, Romero, and
  Gitter]{gelman2021neural}
Sam Gelman, Sarah~A Fahlberg, Pete Heinzelman, Philip~A Romero, and Anthony
  Gitter.
\newblock Neural networks to learn protein sequence-function relationships from
  deep mutational scanning data.
\newblock \emph{bioRxiv}, pages 2020--10, 2021.

\bibitem[Gustafsson et~al.(2020)Gustafsson, Danelljan, and
  Schon]{gustafsson2020evaluating}
Fredrik~K Gustafsson, Martin Danelljan, and Thomas~B Schon.
\newblock Evaluating scalable bayesian deep learning methods for robust
  computer vision.
\newblock In \emph{Proceedings of the IEEE/CVF Conference on Computer Vision
  and Pattern Recognition Workshops}, pages 318--319, 2020.

\bibitem[Hafner et~al.(2020)Hafner, Tran, Lillicrap, Irpan, and
  Davidson]{hafner2020priors}
Danijar Hafner, Dustin Tran, Timothy Lillicrap, Alex Irpan, and James Davidson.
\newblock Noise contrastive priors for functional uncertainty.
\newblock In Ryan~P. Adams and Vibhav Gogate, editors, \emph{Proceedings of The
  35th Uncertainty in Artificial Intelligence Conference}, volume 115 of
  \emph{Proceedings of Machine Learning Research}, pages 905--914. PMLR, 22--25
  Jul 2020.

\bibitem[Hill and Young(2010)]{hill2010getting}
Alan~P Hill and Robert~J Young.
\newblock Getting physical in drug discovery: a contemporary perspective on
  solubility and hydrophobicity.
\newblock \emph{Drug discovery today}, 15\penalty0 (15-16):\penalty0 648--655,
  2010.

\bibitem[Hopf et~al.(2017)Hopf, Ingraham, Poelwijk, Sch{\"a}rfe, Springer,
  Sander, and Marks]{hopf2017mutation}
Thomas~A Hopf, John~B Ingraham, Frank~J Poelwijk, Charlotta~PI Sch{\"a}rfe,
  Michael Springer, Chris Sander, and Debora~S Marks.
\newblock Mutation effects predicted from sequence co-variation.
\newblock \emph{Nature Biotechnology}, 35\penalty0 (2):\penalty0 128--135,
  2017.

\bibitem[Hsu et~al.(2022)Hsu, Nisonoff, Fannjiang, and Listgarten]{hsu2022nbt}
Chloe Hsu, Hunter Nisonoff, Clara Fannjiang, and Jennifer Listgarten.
\newblock Learning protein fitness models from evolutionary and assay-labeled
  data.
\newblock \emph{Nature Biotechnology}, 2022.

\bibitem[Huang et~al.(2021)Huang, Fu, Gao, Zhao, Roohani, Leskovec, Coley,
  Xiao, Sun, and Zitnik]{huang2021therapeutics}
Kexin Huang, Tianfan Fu, Wenhao Gao, Yue Zhao, Yusuf Roohani, Jure Leskovec,
  Connor~W Coley, Cao Xiao, Jimeng Sun, and Marinka Zitnik.
\newblock Therapeutics data commons: machine learning datasets and tasks for
  therapeutics.
\newblock \emph{arXiv preprint arXiv:2102.09548}, 2021.

\bibitem[Izmailov et~al.(2021)Izmailov, Vikram, Hoffman, and
  Wilson]{pmlr-v139-izmailov21a}
Pavel Izmailov, Sharad Vikram, Matthew~D Hoffman, and Andrew Gordon~Gordon
  Wilson.
\newblock What are bayesian neural network posteriors really like?
\newblock In Marina Meila and Tong Zhang, editors, \emph{Proceedings of the
  38th International Conference on Machine Learning}, volume 139 of
  \emph{Proceedings of Machine Learning Research}, pages 4629--4640. PMLR,
  18--24 Jul 2021.
\newblock URL \url{https://proceedings.mlr.press/v139/izmailov21a.html}.

\bibitem[Kauffman and Weinberger(1989)]{kauffman1989nk}
Stuart~A Kauffman and Edward~D Weinberger.
\newblock The nk model of rugged fitness landscapes and its application to
  maturation of the immune response.
\newblock \emph{Journal of theoretical biology}, 141\penalty0 (2):\penalty0
  211--245, 1989.

\bibitem[Kingma and Ba(2015)]{DBLP:journals/corr/KingmaB14}
Diederik~P. Kingma and Jimmy Ba.
\newblock Adam: {A} method for stochastic optimization.
\newblock In Yoshua Bengio and Yann LeCun, editors, \emph{3rd International
  Conference on Learning Representations, {ICLR} 2015, San Diego, CA, USA, May
  7-9, 2015, Conference Track Proceedings}, 2015.
\newblock URL \url{http://arxiv.org/abs/1412.6980}.

\bibitem[Kristiadi et~al.(2020)Kristiadi, Hein, and Hennig]{kristiadi2020being}
Agustinus Kristiadi, Matthias Hein, and Philipp Hennig.
\newblock Being bayesian, even just a bit, fixes overconfidence in relu
  networks.
\newblock In \emph{International Conference on Machine Learning}, pages
  5436--5446. PMLR, 2020.

\bibitem[Lakshminarayanan et~al.(2016)Lakshminarayanan, Pritzel, and
  Blundell]{lakshminarayanan2016simple}
Balaji Lakshminarayanan, Alexander Pritzel, and Charles Blundell.
\newblock Simple and scalable predictive uncertainty estimation using deep
  ensembles.
\newblock \emph{arXiv preprint arXiv:1612.01474}, 2016.

\bibitem[Madani et~al.(2021)Madani, Krause, Greene, Subramanian, Mohr, Holton,
  Olmos, Xiong, Sun, Socher, et~al.]{madani2021deep}
Ali Madani, Ben Krause, Eric~R Greene, Subu Subramanian, Benjamin~P Mohr,
  James~M Holton, Jose~Luis Olmos, Caiming Xiong, Zachary~Z Sun, Richard
  Socher, et~al.
\newblock Deep neural language modeling enables functional protein generation
  across families.
\newblock \emph{bioRxiv}, 2021.

\bibitem[Matsubara et~al.(2021)Matsubara, Oates, and
  Briol]{matsubara2021ridgelet}
Takuo Matsubara, Chris~J. Oates, and FranÃ§ois-Xavier Briol.
\newblock The ridgelet prior: A covariance function approach to prior
  specification for bayesian neural networks.
\newblock \emph{Journal of Machine Learning Research}, 22\penalty0
  (157):\penalty0 1--57, 2021.

\bibitem[Muralidhar et~al.(2018)Muralidhar, Islam, Marwah, Karpatne, and
  Ramakrishnan]{muralidhar2018incorp}
Nikhil Muralidhar, Mohammad~Raihanul Islam, Manish Marwah, Anuj Karpatne, and
  Naren Ramakrishnan.
\newblock Incorporating prior domain knowledge into deep neural networks.
\newblock In \emph{2018 IEEE International Conference on Big Data (Big Data)},
  pages 36--45, 2018.
\newblock \doi{10.1109/BigData.2018.8621955}.

\bibitem[Naesseth et~al.(2020)Naesseth, Lindsten, and
  Blei]{naesseth2020inclusive}
Christian Naesseth, Fredrik Lindsten, and David Blei.
\newblock Markovian score climbing: Variational inference with $kl(p\vert \vert
  q)$.
\newblock In H.~Larochelle, M.~Ranzato, R.~Hadsell, M.~F. Balcan, and H.~Lin,
  editors, \emph{Advances in Neural Information Processing Systems}, volume~33,
  pages 15499--15510. Curran Associates, Inc., 2020.

\bibitem[Ober et~al.(2021)Ober, Rasmussen, and van~der Wilk]{ober2021promises}
Sebastian~W Ober, Carl~E Rasmussen, and Mark van~der Wilk.
\newblock The promises and pitfalls of deep kernel learning.
\newblock \emph{arXiv preprint arXiv:2102.12108}, 2021.

\bibitem[Osband et~al.(2018)Osband, Aslanides, and
  Cassirer]{osband2018randomized}
Ian Osband, John Aslanides, and Albin Cassirer.
\newblock Randomized prior functions for deep reinforcement learning.
\newblock \emph{Advances in Neural Information Processing Systems}, 31, 2018.

\bibitem[Pearce et~al.(2020)Pearce, Leibfried, and
  Brintrup]{pearce2020uncertainty}
Tim Pearce, Felix Leibfried, and Alexandra Brintrup.
\newblock Uncertainty in neural networks: Approximately bayesian ensembling.
\newblock In \emph{International conference on artificial intelligence and
  statistics}, pages 234--244. PMLR, 2020.

\bibitem[Ramsundar et~al.(2019)Ramsundar, Eastman, Walters, Pande, Leswing, and
  Wu]{Ramsundar-et-al-2019}
Bharath Ramsundar, Peter Eastman, Patrick Walters, Vijay Pande, Karl Leswing,
  and Zhenqin Wu.
\newblock \emph{Deep Learning for the Life Sciences}.
\newblock O'Reilly Media, 2019.
\newblock
  \url{https://www.amazon.com/Deep-Learning-Life-Sciences-Microscopy/dp/1492039837}.

\bibitem[Rasmussen and Williams(2006)]{rasmussen2006GP}
Carl~Edward Rasmussen and Christopher K.~I. Williams.
\newblock \emph{Gaussian processes for machine learning.}
\newblock Adaptive computation and machine learning. MIT Press, 2006.
\newblock ISBN 026218253X.

\bibitem[Riesselman et~al.(2018)Riesselman, Ingraham, and
  Marks]{riesselman2018deep}
Adam~J Riesselman, John~B Ingraham, and Debora~S Marks.
\newblock Deep generative models of genetic variation capture the effects of
  mutations.
\newblock \emph{Nature Methods}, 15\penalty0 (10):\penalty0 816--822, 2018.

\bibitem[Romero et~al.(2013)Romero, Krause, and Arnold]{romero2013navigating}
Philip~A Romero, Andreas Krause, and Frances~H Arnold.
\newblock Navigating the protein fitness landscape with gaussian processes.
\newblock \emph{Proceedings of the National Academy of Sciences}, 110\penalty0
  (3):\penalty0 E193--E201, 2013.

\bibitem[Russ et~al.(2020)Russ, Figliuzzi, Stocker, Barrat-Charlaix, Socolich,
  Kast, Hilvert, Monasson, Cocco, Weigt, et~al.]{russ2020evolution}
William~P Russ, Matteo Figliuzzi, Christian Stocker, Pierre Barrat-Charlaix,
  Michael Socolich, Peter Kast, Donald Hilvert, Remi Monasson, Simona Cocco,
  Martin Weigt, et~al.
\newblock An evolution-based model for designing chorismate mutase enzymes.
\newblock \emph{Science}, 369\penalty0 (6502):\penalty0 440--445, 2020.

\bibitem[Sarkisyan et~al.(2016)Sarkisyan, Bolotin, Meer, Usmanova, Mishin,
  Sharonov, Ivankov, Bozhanova, Baranov, Soylemez, et~al.]{sarkisyan2016local}
Karen~S Sarkisyan, Dmitry~A Bolotin, Margarita~V Meer, Dinara~R Usmanova,
  Alexander~S Mishin, George~V Sharonov, Dmitry~N Ivankov, Nina~G Bozhanova,
  Mikhail~S Baranov, Onuralp Soylemez, et~al.
\newblock Local fitness landscape of the green fluorescent protein.
\newblock \emph{Nature}, 533\penalty0 (7603):\penalty0 397--401, 2016.

\bibitem[Sill(1998)]{sill1998monotonic}
Joseph Sill.
\newblock Monotonic networks.
\newblock In M.~Jordan, M.~Kearns, and S.~Solla, editors, \emph{Advances in
  Neural Information Processing Systems}, volume~10. MIT Press, 1998.

\bibitem[Snoek et~al.(2015)Snoek, Rippel, Swersky, Kiros, Satish, Sundaram,
  Patwary, Prabhat, and Adams]{snoek2015scalable}
Jasper Snoek, Oren Rippel, Kevin Swersky, Ryan Kiros, Nadathur Satish,
  Narayanan Sundaram, Mostofa Patwary, Mr~Prabhat, and Ryan Adams.
\newblock Scalable bayesian optimization using deep neural networks.
\newblock In \emph{International conference on machine learning}, pages
  2171--2180. PMLR, 2015.

\bibitem[Sun et~al.(2019)Sun, Zhang, Shi, and Grosse]{sun2018functional}
Shengyang Sun, Guodong Zhang, Jiaxin Shi, and Roger Grosse.
\newblock {FUNCTIONAL} {VARIATIONAL} {BAYESIAN} {NEURAL} {NETWORKS}.
\newblock In \emph{International Conference on Learning Representations}, 2019.

\bibitem[Walters(2021)]{Walters}
Patrick Walters.
\newblock Sfi.
\newblock \url{https://github.com/PatWalters/sfi}, 2021.

\bibitem[Wilson and Izmailov(2020)]{wilson2020bayesian}
Andrew~Gordon Wilson and Pavel Izmailov.
\newblock Bayesian deep learning and a probabilistic perspective of
  generalization.
\newblock \emph{arXiv preprint arXiv:2002.08791}, 2020.

\bibitem[Wilson et~al.(2016)Wilson, Hu, Salakhutdinov, and
  Xing]{wilson2016deep}
Andrew~Gordon Wilson, Zhiting Hu, Ruslan Salakhutdinov, and Eric~P Xing.
\newblock Deep kernel learning.
\newblock In \emph{Artificial intelligence and statistics}, pages 370--378.
  PMLR, 2016.

\bibitem[Wittmann et~al.(2021)Wittmann, Yue, and Arnold]{wittmann2021informed}
Bruce~J Wittmann, Yisong Yue, and Frances~H Arnold.
\newblock Informed training set design enables efficient machine
  learning-assisted directed protein evolution.
\newblock \emph{Cell Systems}, 12\penalty0 (11):\penalty0 1026--1045, 2021.

\bibitem[Wu et~al.(2016)Wu, Dai, Olson, Lloyd-Smith, and Sun]{wu2016adaptation}
Nicholas~C Wu, Lei Dai, C~Anders Olson, James~O Lloyd-Smith, and Ren Sun.
\newblock Adaptation in protein fitness landscapes is facilitated by indirect
  paths.
\newblock \emph{Elife}, 5:\penalty0 e16965, 2016.

\bibitem[Yang et~al.(2020)Yang, Lorch, Graule, Lakkaraju, and
  Doshi-Velez]{NEURIPS2020_95c7dfc5}
Wanqian Yang, Lars Lorch, Moritz Graule, Himabindu Lakkaraju, and Finale
  Doshi-Velez.
\newblock Incorporating interpretable output constraints in bayesian neural
  networks.
\newblock In H.~Larochelle, M.~Ranzato, R.~Hadsell, M.F. Balcan, and H.~Lin,
  editors, \emph{Advances in Neural Information Processing Systems}, volume~33,
  pages 12721--12731. Curran Associates, Inc., 2020.
\newblock URL
  \url{https://proceedings.neurips.cc/paper/2020/file/95c7dfc5538e1ce71301cf92a9a96bd0-Paper.pdf}.

\end{thebibliography}
\bibliographystyle{plainnat}

\section{Appendix}

\subsection{Architecture and optimization hyperparameters}
\label{sec:gridsearch}

For the avGFP and GB1 data sets, cross-validation was used to select the model architecture and optimization hyperparameters. We searched over fully-connected  model architectures with either $1$ or $2$ hidden layers, with the dimensions of the hidden layers varying between, $100$, $200$, and $300$ dimensions. In addition, we searched over the choice of weight-decay used with the Adam optimizer with options of $0.01$, $0.0001$, and $0.000001$.

\subsection{Incorporating Full Covariance Matrices}
It is difficult to use a non-diagonal covariance matrix for the approximate BNN posterior, $q_{_\text{BNN}}(f)$, due to the need to compute a closed-form product of two GPs (or multivariate Gaussians in the finite-dimensional case). Computing the normalized product requires inverting the covariance matrices of the two GPs, which is intractable both for domains with an infinite number of points and for domains with a finite but large set of points (\eg all protein sequences of length L). In order to model correlations between points, an approximation must be made. One approximation is to restrict the domain over which you are defining the function that you are trying to estimate to a small and finite number of points. For example, one can restrict the domain to the set of training and test points.

\subsection{Statistical Significance}
\begin{table*}[h!]
\caption{GB1 task p-values from the Wilcoxon signed-rank test, where for each method, the null hypothesis is that paired samples from \ourmodel\ (\rosettaprior) and the other method come from the same distribution. The alternative hypothesis is that \ourmodel\ (\rosettaprior) improves upon the other method. Each paired sample is computed on a unique test set generated from a random train/val/test split.}
\label{table:TODO}
\begin{center}
\begin{small}
\begin{sc}
\begin{tabular}{lrrr}
\toprule
                     Method &  p-value (LL) &  p-value (RMSE) &  p-value (MAE) \\
\midrule
                        BNN &         0.001 &           0.001 &          0.001 \\
                         NN &         0.001 &           0.001 &          0.001 \\
Stacking: BNN+\rosettaprior &         0.001 &           0.001 &          0.001 \\
  Stacking: BNN+\constprior &         0.001 &           0.001 &          0.001 \\
   \ourmodel\ (\constprior) &         0.022 &           0.022 &          0.022 \\
\bottomrule
\end{tabular}
\end{sc}
\end{small}
\end{center}
\vskip -0.1in
\end{table*}

\begin{table*}[h!]
\caption{GB1 task p-values from the Wilcoxon signed-rank test, where for each method, the null hypothesis is that paired samples from \ourmodel\ (\constprior) and the other method come from the same distribution. The alternative hypothesis is that \ourmodel\ (\constprior) improves upon the other method. Each paired sample is computed on a unique test set generated from a random train/val/test split.}
\label{table:TODO}
\begin{center}
\begin{small}
\begin{sc}
\begin{tabular}{lrrr}
\toprule
                     Method &  p-value (LL) &  p-value (RMSE) &  p-value (MAE) \\
\midrule
                        BNN &         0.001 &           0.001 &          0.001 \\
                         NN &         0.002 &           0.002 &          0.002 \\
Stacking: BNN+\rosettaprior &         0.001 &           0.001 &          0.001 \\
  Stacking: BNN+\constprior &         0.001 &           0.001 &          0.001 \\
 \ourmodel\ (\rosettaprior) &         0.978 &           0.978 &          0.978 \\
\bottomrule
\end{tabular}
\end{sc}
\end{small}
\end{center}
\vskip -0.1in
\end{table*}

\begin{table*}[h!]
\caption{GFP task p-values from the Wilcoxon signed-rank test, where for each method, the null hypothesis is that paired samples from \ourmodel\ (\rosettaprior) and the other method come from the same distribution. The alternative hypothesis is that \ourmodel\ (\rosettaprior) improves upon the other method. Each paired sample is computed on a unique test set generated from a random train/val/test split.}
\label{table:TODO}
\begin{center}
\begin{small}
\begin{sc}
\begin{tabular}{lrrr}
\toprule
                     Method &  p-value (LL) &  p-value (RMSE) &  p-value (MAE) \\
\midrule
                        BNN &         0.001 &           0.001 &          0.001 \\
                         NN &         0.001 &           0.001 &          0.001 \\
Stacking: BNN+\rosettaprior &         0.001 &           0.001 &          0.001 \\
  Stacking: BNN+\constprior &         0.001 &           0.001 &          0.001 \\
   \ourmodel\ (\constprior) &         0.001 &           0.001 &          0.001 \\
\bottomrule
\end{tabular}
\end{sc}
\end{small}
\end{center}
\vskip -0.1in
\end{table*}

\begin{table*}[h!]
\caption{GFP task p-values from the Wilcoxon signed-rank test, where for each method, the null hypothesis is that paired samples from \ourmodel\ (\constprior) and the other method come from the same distribution. The alternative hypothesis is that \ourmodel\ (\constprior) improves upon the other method. Each paired sample is computed on a unique test set generated from a random train/val/test split.}
\label{table:TODO}
\begin{center}
\begin{small}
\begin{sc}
\begin{tabular}{lrrr}
\toprule
                     Method &  p-value (LL) &  p-value (RMSE) &  p-value (MAE) \\
\midrule
                        BNN &         0.001 &           0.001 &          0.001 \\
                         NN &         0.001 &           0.001 &          0.010 \\
Stacking: BNN+\rosettaprior &         0.001 &           0.001 &          0.001 \\
  Stacking: BNN+\constprior &         0.001 &           0.001 &          0.001 \\
 \ourmodel\ (\rosettaprior) &         1.000 &           1.000 &          1.000 \\
\bottomrule
\end{tabular}
\end{sc}
\end{small}
\end{center}
\vskip -0.1in
\end{table*}

\begin{table*}[h!]
\caption{SFI task p-values from the Wilcoxon signed-rank test, where for each method, the null hypothesis is that paired samples from \ourmodel\ (\sfiprior) and the other method come from the same distribution. The alternative hypothesis is that \ourmodel\ (\sfiprior) improves upon the other method. Each paired sample is computed on a unique test set generated from a random train/val/test split.}
\label{table:TODO}
\begin{center}
\begin{small}
\begin{sc}
\begin{tabular}{lrrr}
\toprule
                 Method &  p-value (LL) &  p-value (RMSE) &  p-value (MAE) \\
\midrule
                    BNN &         0.002 &           0.001 &          0.001 \\
                     NN &         0.001 &           0.001 &          0.001 \\
Stacking: BNN+\sfiprior &         0.001 &           1.000 &          0.246 \\
\bottomrule
\end{tabular}
\end{sc}
\end{small}
\end{center}
\vskip -0.1in
\end{table*}

\end{document}